\documentclass[10pt,conference,letterpaper]{manuscript}
\usepackage{times,amsmath,epsfig}

\usepackage{amssymb}
\usepackage{graphicx}
\usepackage{amsmath}
\usepackage{subfigure}
\usepackage{cite}
\usepackage{hyperref}
\usepackage{multirow}  
\usepackage{algorithmic}
\usepackage{algorithm}
\usepackage{balance}
\newcommand{\tabincell}[2]{\begin{tabular}{@{}#1@{}}#2\end{tabular}} 

\title{Drug-drug Interaction Extraction via Recurrent Neural Network with Multiple Attention Layers}
\author{%
{Zibo Yi, Shasha Li, Jie Yu, Qingbo Wu }%
\vspace{1.6mm}\\
\fontsize{10}{10}\selectfont\itshape
$~^{}$College of Computer, National University of Defense Technology\\
Changsha, Hunan, China\\
\fontsize{9}{9}\selectfont\ttfamily\upshape
$~^{}$\{yizibo14, shashali, yj, qingbo.wu\}@nudt.edu.cn\\
}
\begin{document}
\maketitle
\begin{abstract} 
Drug-drug interaction (DDI) is a vital information when physicians and pharmacists intend to co-administer two or more drugs. Thus, several DDI databases are constructed to avoid mistakenly combined use. In recent years, automatically extracting DDIs from biomedical text has drawn researchers' attention. However, the existing work utilize either complex feature engineering or NLP tools, both of which are insufficient for sentence comprehension. Inspired by the deep learning approaches in natural language processing, we propose a recurrent neural network model with multiple attention layers for DDI classification. We evaluate our model on 2013 SemEval DDIExtraction dataset. The experiments show that our model classifies most of the drug pairs into correct DDI categories, which outperforms the existing NLP or deep learning methods.
\end{abstract}

% NOTE keywords are not used for conference papers so do not populate them
\begin{keywords}
Drug-drug interaction, Recurrent neural network, Attention layer, Deep learning, Natural language processing
\end{keywords}
\section{Introduction}

Drug-drug interaction (DDI) is a situation when one drug increases or decreases the effect of another drug \cite{Tari2010Discovering}. Adverse drug reactions may cause severe side effect, if two or more medicines were taken and their DDI were not investigated in detail. DDI is a common cause of illness, even a cause of death \cite{lazarou1998incidence}. Thus, DDI databases for clinical medication decisions are proposed by some researchers. These databases such as SFINX\cite{bottiger2009sfinx}, KEGG \cite{takarabe2011network-based}, CredibleMeds \cite{Shankar2014CredibleMeds} help physicians and pharmacists avoid most adverse drug reactions. 

Traditional DDI databases are manually constructed according to clinical records, scientific research and drug specifications. For instance, The sentence ``With combined use, clinicians should be aware, when \textbf{phenytoin} is added, of the potential for reexacerbation of pulmonary symptomatology due to lowered serum \textbf{theophylline} concentrations \cite{Sklar1985Enhanced}'', which is from a pharmacotherapy report, describe the side effect of phenytoin and theophylline's combined use. Then this information on specific medicines will be added to DDI databases. As drug-drug interactions have being increasingly found, manually constructing DDI database would consume a lot of manpower and resources. 

There has been many efforts to automatically extract DDIs from natural language \cite{Tari2010Discovering, Lu2015A, Bui2014A, Liu2016Dependency, Liu2016Drug, chowdhury2013fbk, Thomas2013WBI, sahu2017drug}, mainly medical literature and clinical records. These works can be divided into the following categories:

\begin{itemize}
\item Text analysis and statistics based approach \cite{Tari2010Discovering, Lu2015A, Melnikov2014Retrieval}. This kind of work utilizes NLP tools to analysis biomedical text's semantics or statistics features (such as TF-IDF) before the DDI decision. However, the semantics and statistics features are insufficient for understanding the whole text. Even worse, NLP toolkits are imperfect and may propagate error to the classification.
\item Feature based machine learning approach \cite{Bui2014A, chowdhury2013exploiting, rastegar2013extraction, chowdhury2013fbk, Thomas2013WBI}. Such method always need complex feature engineering. In addition, the quality of feature engineering have a great effect on the precision of DDI classification, which becomes the shortcoming of such method.
\item Deep learning based approach \cite{Liu2016Dependency, Liu2016Drug, sahu2017drug}. Deep learning neural networks, such as convolutional neural networks (CNN) and long short-term memory networks (LSTM), have been utilized for DDI extraction. Deep learning method avoids complicated feature engineering since CNN and LSTM can extract semantics features automatically through a well designed network.
\end{itemize}

To avoid complex feature engineering and NLP toolkits' usage, we employ deep learning approaches for sentence comprehension as a whole. Our model takes in a sentence from biomedical literature which contains a drug pair and outputs what kind of DDI this drug pair belongs. This assists physicians refrain from improper combined use of drugs. In addition, the word and sentence level attentions are introduced to our model for better DDI predictions.

\begin{figure}
\centering  
\includegraphics [height=1.7in]{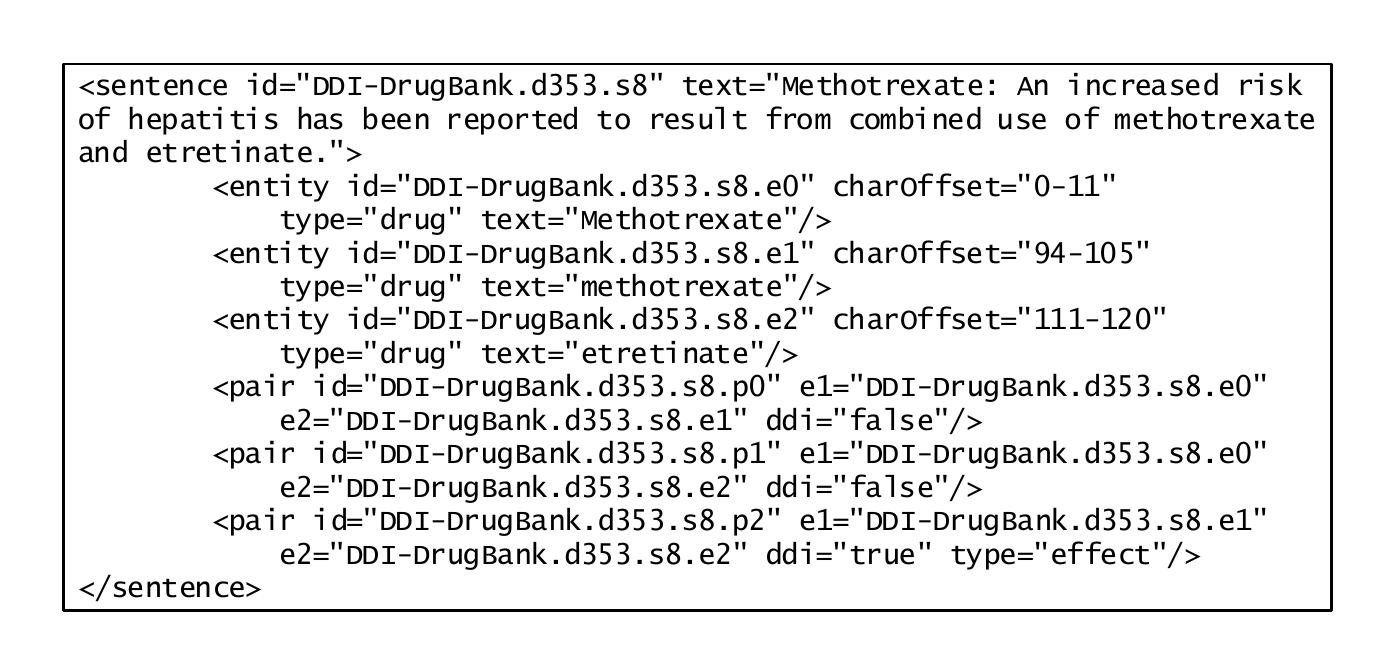} 
\caption{Partial records in DDI corpus} 
\label{fig:record} 
\end{figure}

We train our language comprehension model with labeled instances. Figure \ref{fig:record} shows partial records in DDI corpus \cite{Herrero2013The}. We extract the sentence and drug pairs in the records. There are 3 drug pairs in this example thus we have 3 instances. The DDI corpus annotate each drug pair in the sentence with a DDI type. The DDI type, which is the most concerned information, is described in table \ref{tab:instances}. The details about how we train our model and extract the DDI type from text are described in the remaining sections.

\begin{table*}[ht]
\caption{The DDI types and corresponding examples}
\label{tab:instances}
\centering
\begin{tabular}{cccc}
 \hline
DDI types & Definition & Example sentence & Drug pair\\
 \hline
 \hline
 False & \tabincell{c}{An interaction between the two drugs \\is not shown in the sentence.} & \tabincell{c}{Concomitantly given thiazide diuretics\\ did not interfere with the\\ absorption of a tablet of digoxin.} & \tabincell{c}{thiazide diuretics, digoxin}\\
 \hline
 Mechanism & \tabincell{c}{An pharmacokinetic mechanism\\ is shown in the sentence.} & \tabincell{c}{Additional iron significantly inhibited\\ the absorption of cobalt.} & \tabincell{c}{iron, cobalt}\\
 \hline
 Effect & \tabincell{c}{The effect of two drugs' combination\\ use is shown in the sentence.} & \tabincell{c}{Methotrexate: An increased risk of \\hepatitis has been reported to result from\\ combined use of methotrexate and etretinate.} & \tabincell{c}{methotrexate, etretinate}\\
  \hline
 Advise & \tabincell{c}{An advise about two drugs\\ is given in the sentence.} & \tabincell{c}{UROXATRAL should NOT be used in\\ combination with other alpha-blockers.} & \tabincell{c}{UROXATRAL, alpha-blockers}\\
   \hline
 Int & \tabincell{c}{A drug interaction without any further\\ information is mentioned in the sentence.} & \tabincell{c}{Clinical implications of warfarin\\ interactions with five sedatives.} & \tabincell{c}{warfarin, sedatives}\\
 \hline
 \end{tabular}
\end{table*}

\section{Related Work}
In DDI extraction task, NLP methods or machine learning approaches are proposed by most of the work. Chowdhury \cite{chowdhury2013exploiting} and Thomas \emph{et al.} \cite{Thomas2013WBI} proposed methods that use linguistic phenomenons and two-stage SVM to classify DDIs. FBK-irst \cite{chowdhury2013fbk} is a follow-on work which applies kernel method to the existing model and outperforms it. 

Neural network based approaches have been proposed by several works. Liu \emph{et al.} \cite{Liu2016Drug} employ CNN for DDI extraction for the first time which outperforms the traditional machine learning based methods. Limited by the convolutional kernel size, the CNN can only extracted features of continuous 3 to 5 words rather than distant words. Liu \emph{et al.} \cite{Liu2016Dependency} proposed dependency-based CNN to handle distant but relevant words. Sahu \emph{et al.} \cite{sahu2017drug} proposed LSTM based DDI extraction approach and outperforms CNN based approach, since LSTM handles sentence as a sequence instead of slide windows. To conclude, Neural network based approaches have advantages of 1) less reliance on extra NLP toolkits, 2) simpler preprocessing procedure, 3) better performance than text analysis and machine learning methods.

Drug-drug interaction extraction is a relation extraction task of natural language processing. Relation extraction aims to determine the relation between two given entities in a sentence. In recent years, attention mechanism and various neural networks are applied to relation extraction \cite{Zhou2016Attention, zhang2015bidirectional, zeng2014relation, Wang2016Relation, Nguyen2015Relation}. Convolutional deep neural network are utilized for extracting sentence level features in \cite{zeng2014relation}. Then the sentence level features are concatenated with lexical level features, which are obtained by NLP toolkit WordNet \cite{Lin1999WordNet}, followed by a multilayer perceptron (MLP) to classify the entities' relation. A fixed work is proposed by Nguyen \emph{et al.} \cite{Nguyen2015Relation}. The convolutional kernel is set various size to capture more \emph{n}-gram features. In addition, the word and position embedding are trained automatically instead of keeping constant as in \cite{zeng2014relation}. Wang \emph{et al.} \cite{Wang2016Relation} introduce multi-level attention mechanism to CNN in order to emphasize the keywords and ignore the non-critical words during relation detection. The attention CNN model outperforms previous state-of-the-art methods.

Besides CNN, Recurrent neural network (RNN) has been applied to relation extraction as well. Zhang \emph{et al.} \cite{zhang2015bidirectional} utilize long short-term memory network (LSTM), a typical RNN model, to represent sentence. The bidirectional LSTM chronologically captures the previous and future information, after which a pooling layer and MLP have been set to extract feature and classify the relation. Attention mechanism is added to bidirectional LSTM in \cite{Zhou2016Attention} for relation extraction. An attention layer gives each memory cell a weight so that classifier can catch the principal feature for the relation detection. The Attention based bidirectional LSTM has been proven better than previous work.

\section{Proposed Model}

In this section, we present our bidirectional recurrent neural network with multiple attention layer model. The overview of our architecture is shown in figure \ref{fig:arch}. For a given instance, which describes the details about two or more drugs, the model represents each word as a vector in embedding layer. Then the bidirectional RNN layer generates a sentence matrix, each column vector in which is the semantic representation of the corresponding word. The word level attention layer transforms the sentence matrix to vector representation. Then sentence level attention layer generates final representation for the instance by combining several relevant sentences in view of the fact that these sentences have the same drug pair. Followed by a softmax classifier, the model classifies the drug pair in the given instance as specific DDI.

\subsection{Preprocessing}\label{subsec:processing}
The DDI corpus contains thousands of XML files, each of which are constructed by several records. For a sentence containing $n$ drugs, there are ${\rm C}_n^2$ drug pairs. We replace the interested two drugs with ``drug1'' and ``drug2'' while the other drugs are replaced by ``durg0'', as in \cite{Liu2016Drug} did. This step is called drug blinding. For example, the sentence in figure \ref{fig:record} generates 3 instances after drug blinding: ``drug1: an increased risk of hepatitis has been reported to result from combined use of drug2 and drug0'', ``drug1: an increased risk of hepatitis has been reported to result from combined use of drug0 and drug2'', ``drug0: an increased risk of hepatitis has been reported to result from combined use of drug1 and drug2''. The drug blinded sentences are the instances that are fed to our model.

We put the sentences with the same drug pairs together as a set, since the sentence level attention layer (will be described in Section \ref{subsec:sentence_level}) will use the sentences which contain the same drugs.

\subsection{Embedding Layer}
Given an instance $S=(w_1, w_2, ..., w_t)$ which contains specified two drugs $w_u=``{\rm{drug1}}"$, $w_v=``{\rm{drug2}}"$, each word is embedded in a $d=d_{WE}+2d_{PE}$ dimensional space ($d_{WE}$, $d_{PE}$ are the dimension of word embedding and position embedding). 
The look up table function $LT_{\cdot}(\cdot)$ maps a word or a relative position to a column vector. After embedding layer the sentence is represented by $S=(\textit{\textbf{x}}_1, \textit{\textbf{x}}_2, ..., \textit{\textbf{x}}_t)$, where

\begin{equation}
\label{eq:embedding}
\textit{\textbf{x}}_i=(LT_{W}(w_i)^{\rm T}, (LT_{P}(i-u)^{\rm T}, (LT_{P}(i-v)^{\rm T})^{\rm T}
\end{equation}

The $LT_{\cdot}(\cdot)$ function is usually implemented with matrix-vector product. Let $\overline{w_i}$, $\overline{k}$ denote the one-hot representation (column vector) of word and relative distance. $E_w$, $E_p$ are word and position embedding query matrix. The look up functions are implemented by

\begin{equation}
\label{eq:lookuptable}
LT_{W}(w_i)={E_w}{\overline{w_i}}, LT_{P}(k)={E_p}{\overline{k}}
\end{equation}

Then the word sequence $S=(\textit{\textbf{x}}_1, \textit{\textbf{x}}_2, ..., \textit{\textbf{x}}_t)$ is fed to the RNN layer. Note that the sentence will be filled with $\textbf{0}$ if its length is less than $t$.

\subsection{Bidirectional RNN Encoding Layer}
The words in the sequence are read by RNN's gated recurrent unit (GRU) one by one. The GRU takes the current word $\textit{\textbf{x}}_i$ and the previous GRU's hidden state $h_{i-1}$ as input. The current GRU encodes $h_{i-1}$ and $\textit{\textbf{x}}_i$ into a new hidden state $h_i$ (its dimension is $d_h$, a hyperparameter), which can be regarded as informations the GRU remembered.

\begin{figure}
\centering  
\includegraphics [height=1.7in]{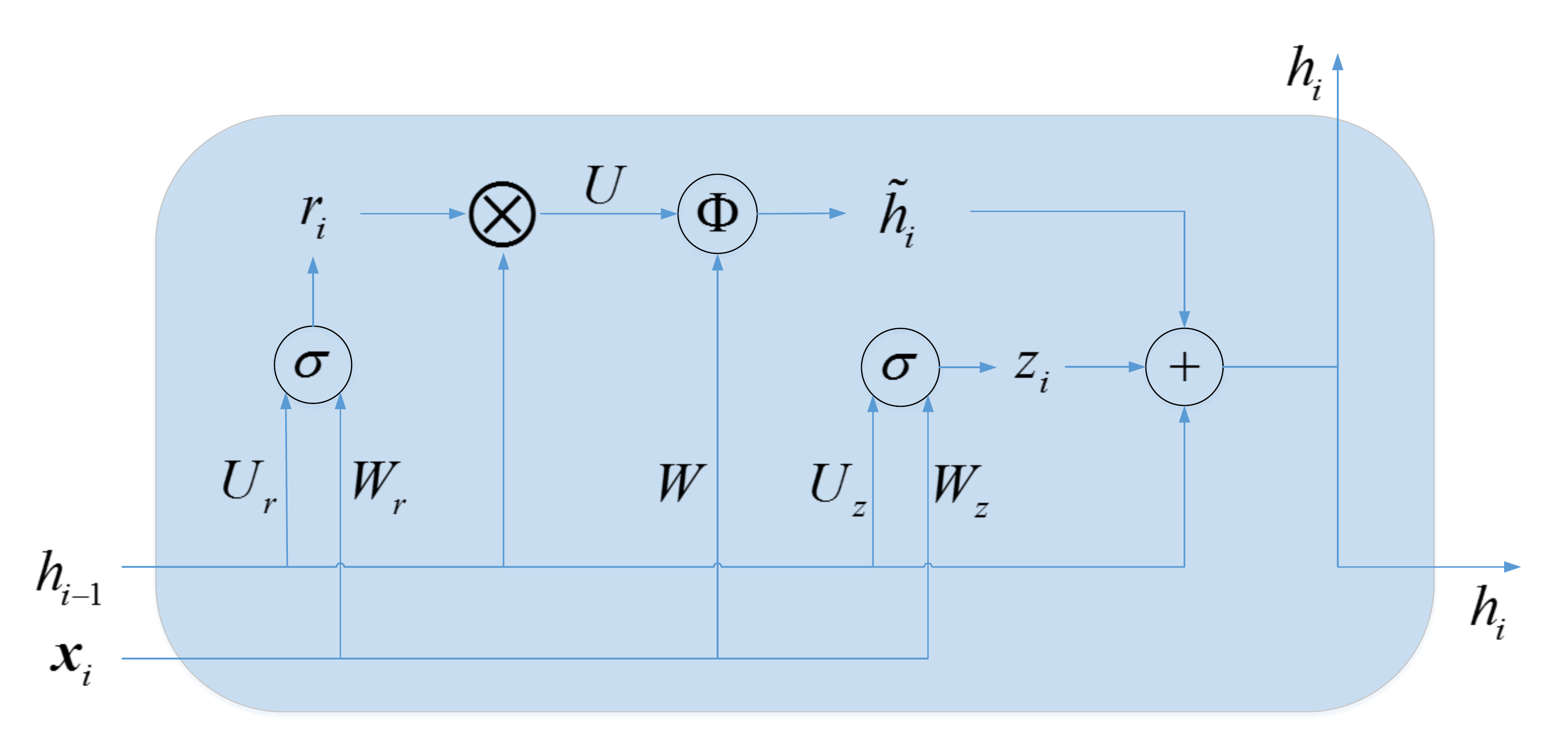} 
\caption{The Gated Recurrent Unit} 
\label{fig:gru} 
\end{figure}

Figure \ref{fig:gru} shows the details in GRU. The reset gate $r_i$ selectively forgets informations delivered by previous GRU. Then the hidden state becomes $\tilde{h}_i$. The update gate $z_i$ updates the informations according to $\tilde{h}_i$ and $h_{i-1}$. The equations below describe these procedures. Note that $\otimes$ stands for element wise multiplication.

\begin{equation}
r_i=\sigma(W_r\textit{\textbf{x}}_i+U_{r}h_{i-1})
\end{equation}

\begin{equation}
\tilde{h}_i={\rm \Phi}(W\textit{\textbf{x}}_i+U(r_{i}\otimes{h_{i-1}}))
\end{equation}

\begin{equation}
z_i=\sigma(W_z\textit{\textbf{x}}_i+U_{z}h_{i-1})
\end{equation}

\begin{equation}
h_i=z_i\otimes{h_{i-1}}+((1, 1, ..., 1)^{\rm T}-z_i)\otimes{\tilde{h}_i}
\end{equation}

\begin{figure}
\centering  
\includegraphics [height=3.2in]{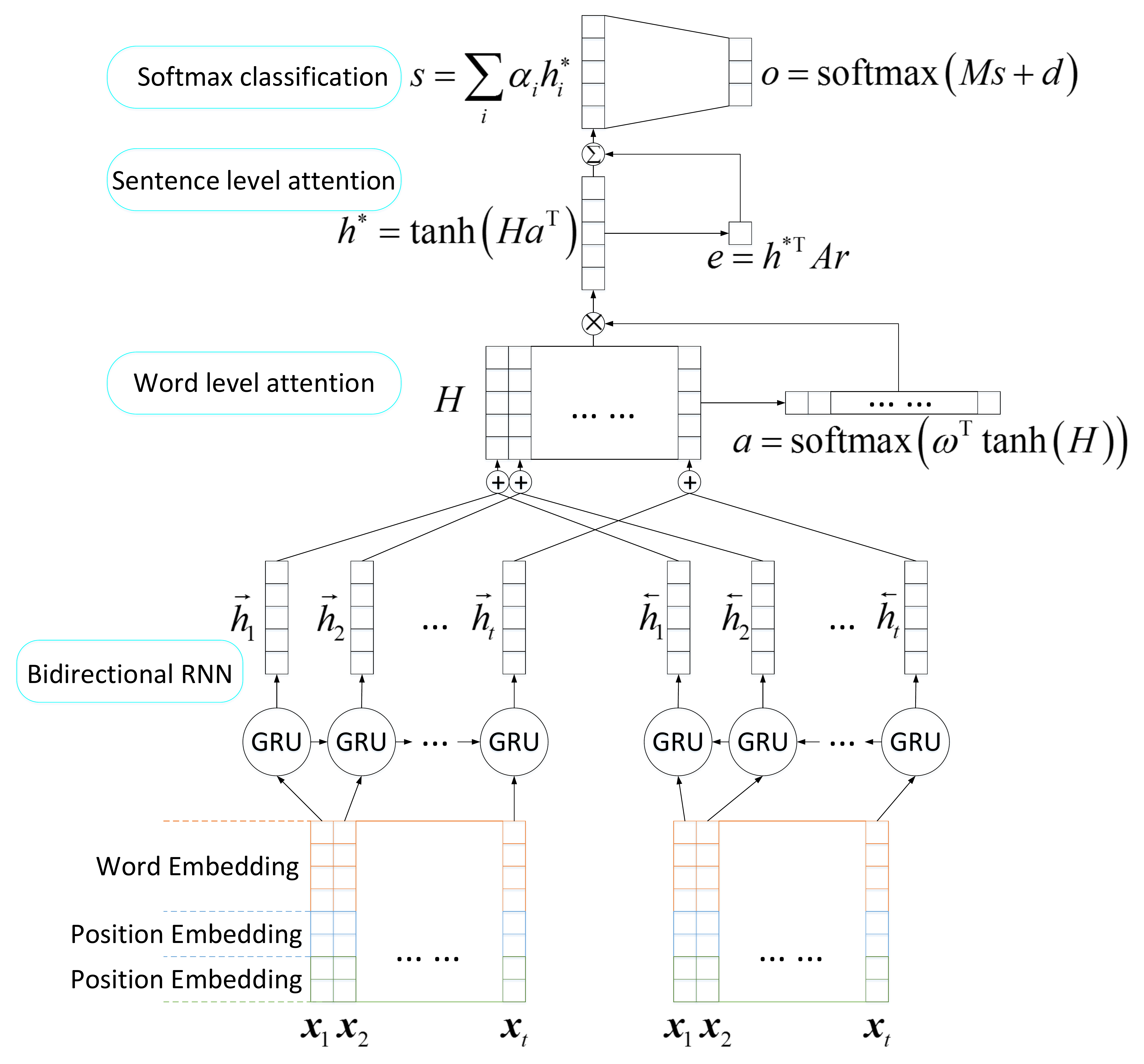} 
\caption{The bidirectional recurrent neural network with multiple attentions} 
\label{fig:arch} 
\end{figure}

The bidirectional RNN contains forward RNN and backward RNN. Forward RNN reads sentence from $\textit{\textbf{x}}_1$ to $\textit{\textbf{x}}_t$, generating $\overrightarrow{h}_1$, $\overrightarrow h_2$, ..., $\overrightarrow h_t$. Backward RNN reads sentence from $\textit{\textbf{x}}_t$ to $\textit{\textbf{x}}_1$, generating $\overleftarrow{h}_t$, $\overleftarrow h_{t-1}$, ..., $\overleftarrow h_1$. Then the encode result of this layer is 

\begin{equation}
H=(\overrightarrow{h}_1+\overleftarrow{h}_1, \overrightarrow{h}_2+\overleftarrow{h}_2, ..., \overrightarrow{h}_t+\overleftarrow{h}_t)
\end{equation}

We apply dropout technique in RNN layer to avoid overfitting. Each GRU have a probability (denoted by $Pr_{dp}$, also a hyperparameter) of being dropped. The dropped GRU has no output and will not affect the subsequent GRUs. With bidirectional RNN and dropout technique, the input $S=(\textit{\textbf{x}}_1, \textit{\textbf{x}}_2, ..., \textit{\textbf{x}}_t)$ is encoded into sentence matrix $H$.

\subsection{Word Level Attention}
The purpose of word level attention layer is to extract sentence representation (also known as feature vector) from encoded matrix. We use word level attention instead of max pooling, since attention mechanism can determine the importance of individual encoded word in each row of $H$. Let $\omega$ denotes the attention vector (column vector), $a$ denotes the filter that gives each element in the row of $H$ a weight. The following equations shows the attention operation, which is also illustrated in figure \ref{fig:arch}.

\begin{equation}
a={\rm{softmax}}(\omega^{\rm T}{\rm{tanh}}(H))
\end{equation}

\begin{equation}
h^*={\rm{tanh}}(Ha^{\rm T})
\end{equation}

The softmax function takes a vector $v=[v_1, v_2, ..., v_n]$ as input and outputs a vector,

\begin{equation}
{\rm{softmax}}(v)=[\frac{e^{v_1}}{\sum_{i=1}^n{e^{v_i}}}, \frac{e^{v_2}}{\sum_{i=1}^n{e^{v_i}}}, ..., \frac{e^{v_n}}{\sum_{i=1}^n{e^{v_i}}}]
\end{equation}

$h^*$ denotes the feature vector captured by this layer. Several approaches \cite{sahu2017drug, Zhou2016Attention} use this vector and softmax classifier for classification. Inspired by \cite{Lin2016Neural} we propose the sentence level attention to combine the information of other sentences for a improved DDI classification.

\subsection{Sentence Level Attention}\label{subsec:sentence_level}

The previous layers captures the features only from the given sentence. However, other sentences may contains informations that contribute to the understanding of this sentence. It is reasonable to look over other relevant instances when determine two drugs' interaction from the given sentence. In our implementation, the instances that have the same drug pair are believed to be relevant. The relevant instances set is denoted by $\mathcal{L} = \{h^*_1, h^*_2, ..., h^*_N\}$, where $h^*_i$ is the sentence feature vector. $e_i$ stands for how well the instance $h_i^*$ matches its DDI $r$ (Vector representation of a specific DDI). $A$ is a diagonal attention matrix, multiplied by which the feature vector $h_i^*$ can concentrate on those most representative features. 

\begin{equation}
e_i=h^{*\rm T}_iAr
\end{equation}

\begin{equation}
\alpha_i=\frac{{\rm{exp}}(e_i)}{\sum_{k=1}^{N}{\rm{exp}}(e_k)}
\end{equation}

$\alpha_i$ is the softmax result of $e_i$. The final sentence representation is decided by all of the relevant sentences' feature vector, as Equation \ref{eq:s} shows.

\begin{equation}
\label{eq:s}
s=\sum_{i=1}^{N}\alpha_ih^*_i
\end{equation}

Note that the set $\mathcal{L}$ is gradually growing as new sentence with the same drugs pairs is found when training. An instance $S=(\textit{\textbf{x}}_1, \textit{\textbf{x}}_2, ..., \textit{\textbf{x}}_t)$ is represented by $h^*$ before sentence level attention. The sentence level attention layer finds the set $\mathcal{L}$, instances in which have the same drug pair as in $S$, and put $S$ in $\mathcal{L}$. Then the final sentence representation $s$ is calculated in this layer.

\subsection{Classification and Training}\label{subsec:cla_train}

A given sentence $S=(w_1, w_2, ..., w_t)$ is finally represented by the feature vector $s$. Then we feed it to a softmax classifier. Let $C$ denotes the set of all kinds of DDI. The output $o\in R^{|C|}$ is the probabilities of each class $S$ belongs.

\begin{equation}
o={\rm{softmax}}(Ms+d)
\end{equation}

We use cross entropy cost function and $L^2$ regularization as the optimization objective. For $i$-th instance, $Y_i$ denotes the one-hot representation of it's label, where the model outputs $o_i$. The cross entropy cost is: 

\begin{equation}
{l}_i=-{\rm{ln}}Y_i^{\rm T}o_i
\end{equation}

For a mini-batch $\mathcal M=\{S_1, S_2, ..., S_M\}$, the optimization objective is:

\begin{equation}
J(\theta)=-\frac{1}{|\mathcal M|}\sum_{i=1}^{|\mathcal M|}{{\rm{ln}}Y_i^{\rm T}o_i} + \lambda ||\theta||_2^2
\end{equation}

All parameters in this model is:

\begin{equation}
\theta=\{E_w, E_p, W_r, U_r, W, U, W_z, U_z, \omega, A, r, M, d\}
\end{equation}

We optimize the parameters of objective function $J(\theta)$ with Adam \cite{kingma2014adam}, which is a variant of mini-batch stochastic gradient descent. During each train step, the gradient of $J(\theta)$ is calculated. Then $\theta$ is adjusted according to the gradient. After the end of training, we have a model that is able to predict two drugs' interactions when a sentence about these drugs is given.

\subsection{DDI Prediction}
The model is trained for DDI classification. The parameters in list $\theta$ are tuned during the training process. Given a new sentence with two drugs, we can use this model to classify the DDI type.

The DDI prediction follows the procedure described in Section \ref{subsec:processing} - \ref{subsec:cla_train}. The given sentence is eventually represented by feature vector $s$. Then $s$ is classified to a specific DDI type with a softmax classifier. In next section, we will evaluate our model's DDI prediction performance and see the advantages and shortcomings of our model.

\begin{figure*}[!t]
\centering
\subfigure {{\includegraphics[height=2.2in,angle=0]{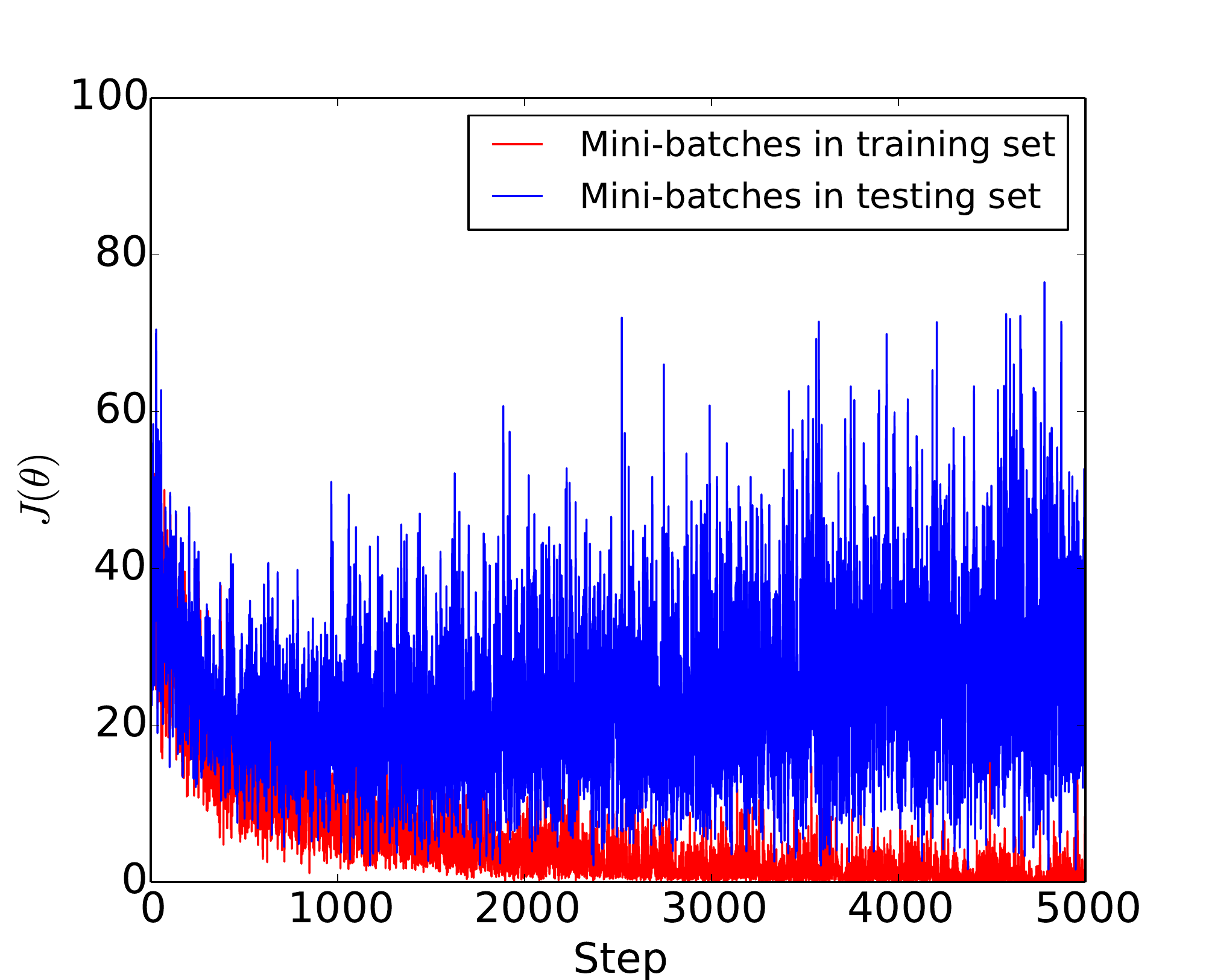}}\label{subfig:train_loss}}
\subfigure {{\includegraphics[height=2.2in,angle=0]{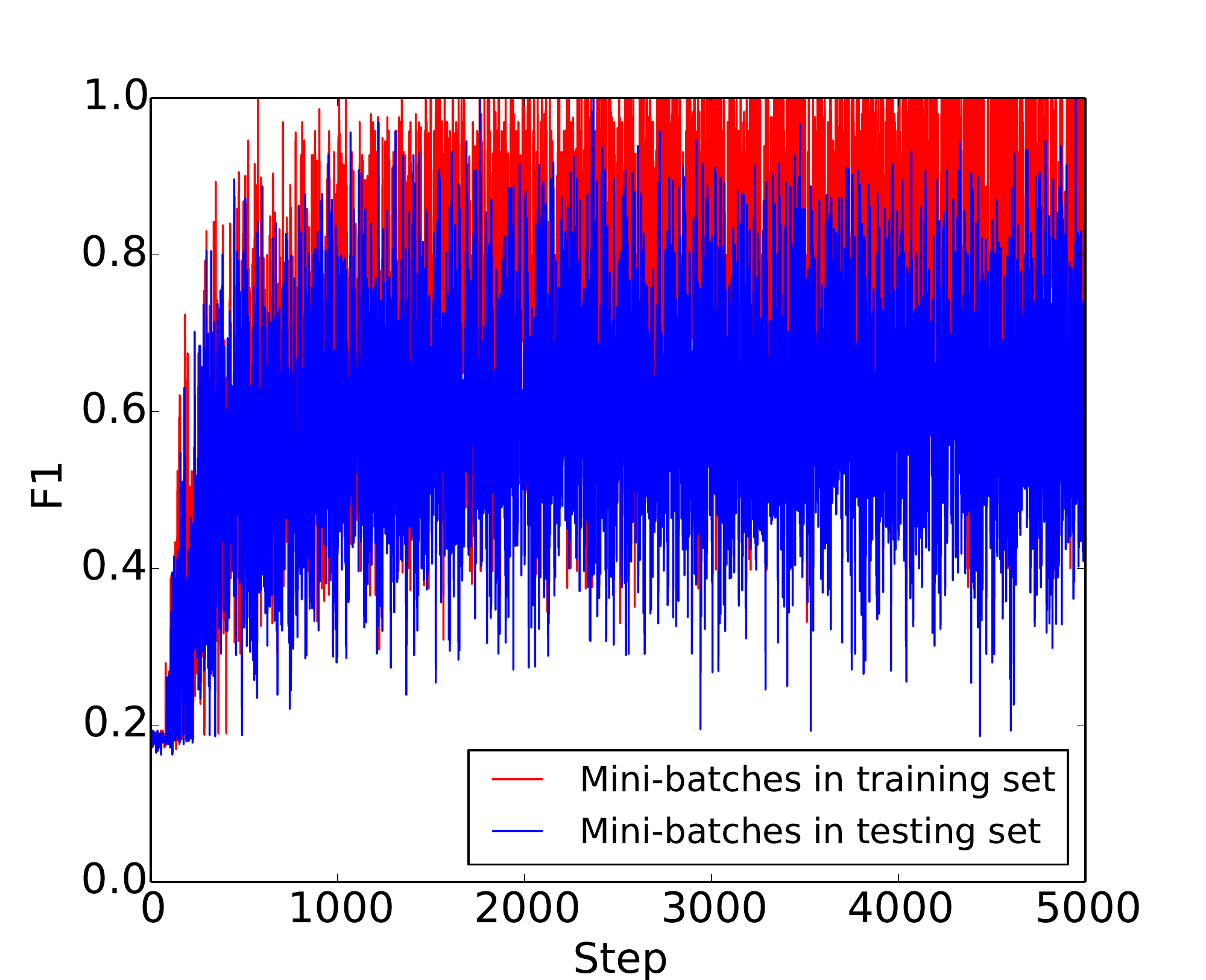}}\label{subfig:train_f1}}
\caption{The objective function and F1 in the train process}
\label{fig:train_process}
\end{figure*}

\section{Experiments}
\subsection {Datasets and Evaluation Metrics}
We use the DDI corpus of the 2013 DDIExtraction challenge \cite{Herrero2013The} to train and test our model. The DDIs in this corpus are classified as five types. We give the definitions of these types and their example sentences, as shown in table \ref{tab:instances}. This standard dataset is made up of training set and testing set. We use the same metrics as in other drug-drug interaction extraction literature \cite{Thomas2013WBI, chowdhury2013fbk, Zhao2016Drug, Liu2016Drug, Liu2016Dependency, sahu2017drug}: the overall precision, recall, and F1 score on testing set. $C$ denotes the set of \{False, Mechanism, Effect, Advise, Int\}. The precision and recall of each $c\in C$ are calculated by

\begin{equation}
P_c=\frac{\#\ DDI\ is\ c\ and\ is\ classified\ as\ c}{\#\ Classified\ as\ c}
\end{equation}

\begin{equation}
R_c=\frac{\#\ DDI\ is\ c\ and\ is\ classified\ as\ c}{\#\ DDI\ is\ c}
\end{equation}

Then the overall precision, recall, and F1 score are calculated by

\begin{equation}
P=\frac{1}{|C|}\sum_{c\in C}{P_c},\ R=\frac{1}{|C|}\sum_{c\in C}{R_c},\ F1=\frac{2PR}{P+R}
\end{equation}

Besides, we evaluate the captured feature vectors with t-SNE \cite{maaten2008visualizing}, a visualizing and intuitive way to map a high dimensional vector into a 2 or 3-dimensional space. If the points in a low dimensional space are easy to be split, the feature vectors are believed to be more distinguishable.

\begin{table*}[ht]
\caption{Performance comparison with other approaches}
\label{tab:performance}
\centering
\begin{tabular}{ccccc}
 \hline
 \multirow{2}{*}{\tabincell{c}{Systems}} & \multirow{2}{*}{Methods} & \multicolumn{3}{c}{Performance} \\
 \cline{3-5}
   ~ & ~ & P & R & F1 \\
 \hline
 \hline
{WBI \cite{Thomas2013WBI}} & Two stage SVM classification & 0.6420 & 0.5790 & 0.6090\\
\hline
{FBK-ist \cite{chowdhury2013fbk}} & Hand crafted features + SVM & 0.6460 & 0.6560 & 0.6510\\
 \hline
 {SCNN \cite{Zhao2016Drug}} & Two stage syntax CNN & 0.725 & 0.651 & 0.686\\
 \hline
 {Liu \emph{et al.} \cite{Liu2016Drug}} & CNN + Pre-trained WE & 0.7572 & 0.6466 & 0.6975\\
 \hline
 {DCNN \cite{Liu2016Dependency}} & Dependency-based CNN + Pretrained WE & \textbf{0.7721} & 0.6435 & 0.7019\\
 \hline
 {Sahu \emph{et al.} \cite{sahu2017drug}} & bidirectional LSTM + ATT & 0.7341 & 0.6966 & 0.7148\\
 \hline
 {This paper} & RNN + dynamic WE + 2ATT & 0.7367 & \textbf{0.7079} & \textbf{0.7220} \\
 \hline
 \end{tabular}
\end{table*}

\begin{figure}
\centering  
\includegraphics [height=2.5in]{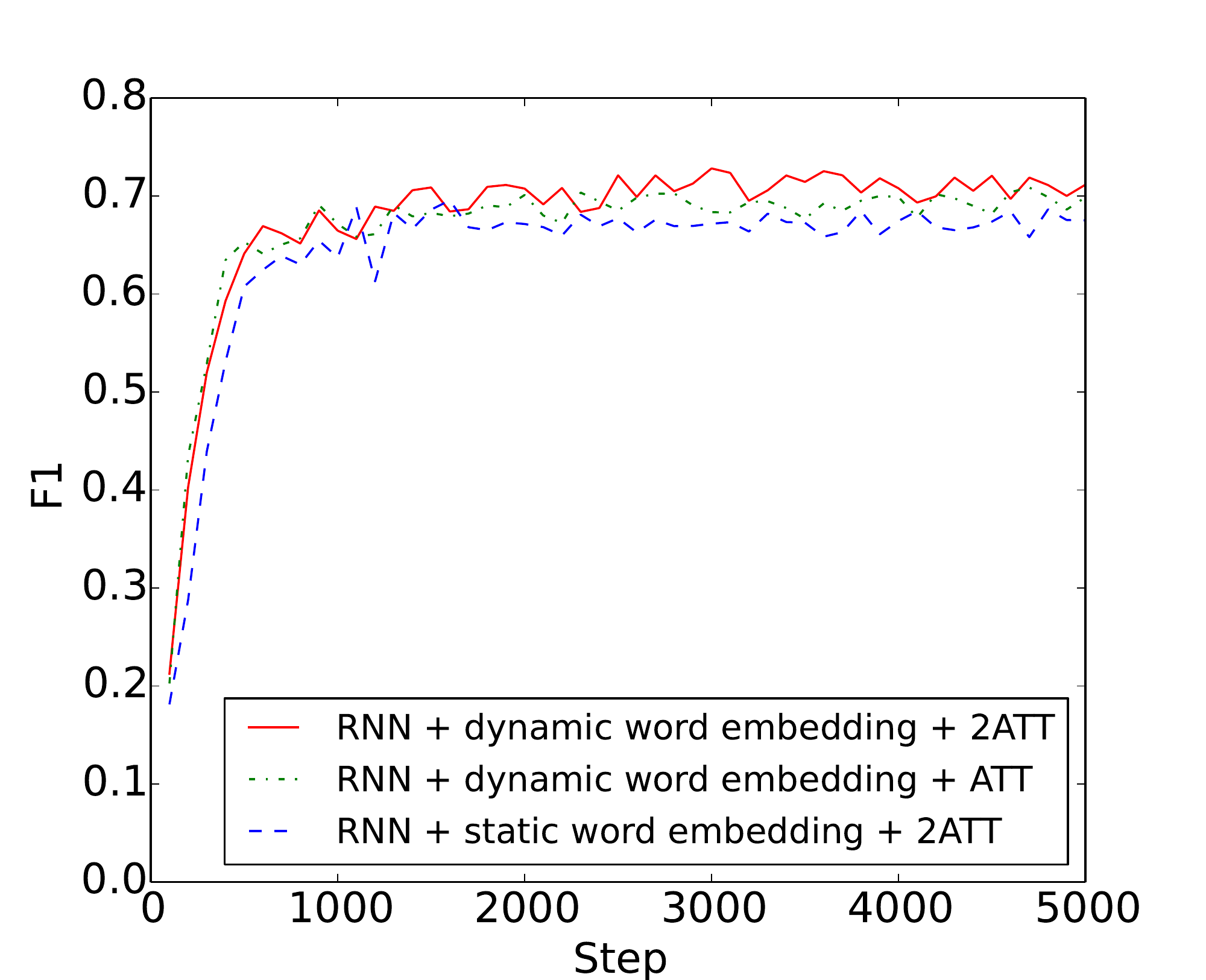} 
\caption{The F1 scores on the whole testing set} 
\label{fig:testf1} 
\end{figure}

\begin{figure*}[!t]
\centering
\subfigure[Static word embedding + 2ATT] {{\includegraphics[height=1.73in,angle=0]{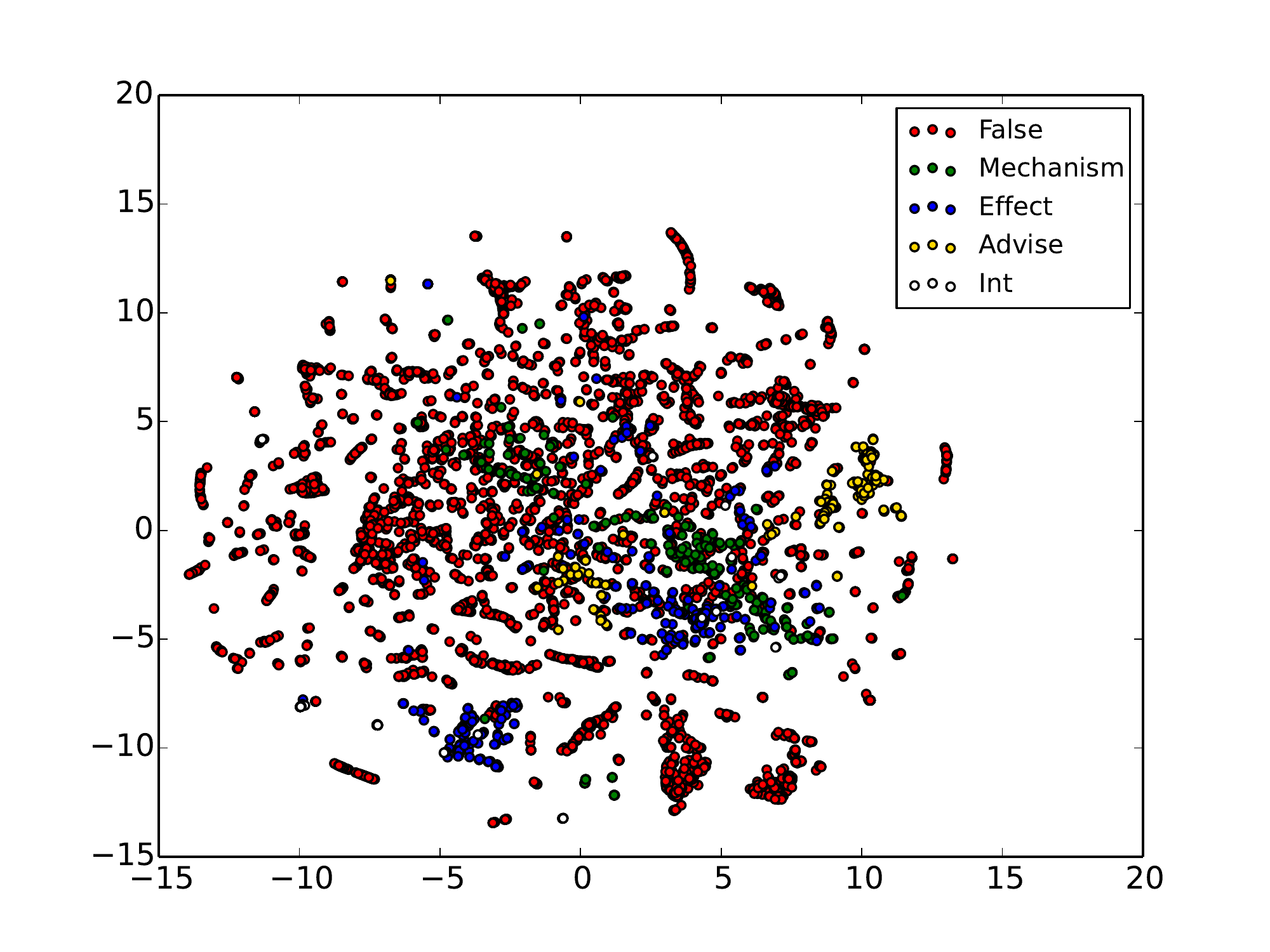}}\label{subfig:features_static}}
\subfigure[Dynamic word embedding + ATT] {{\includegraphics[height=1.73in,angle=0]{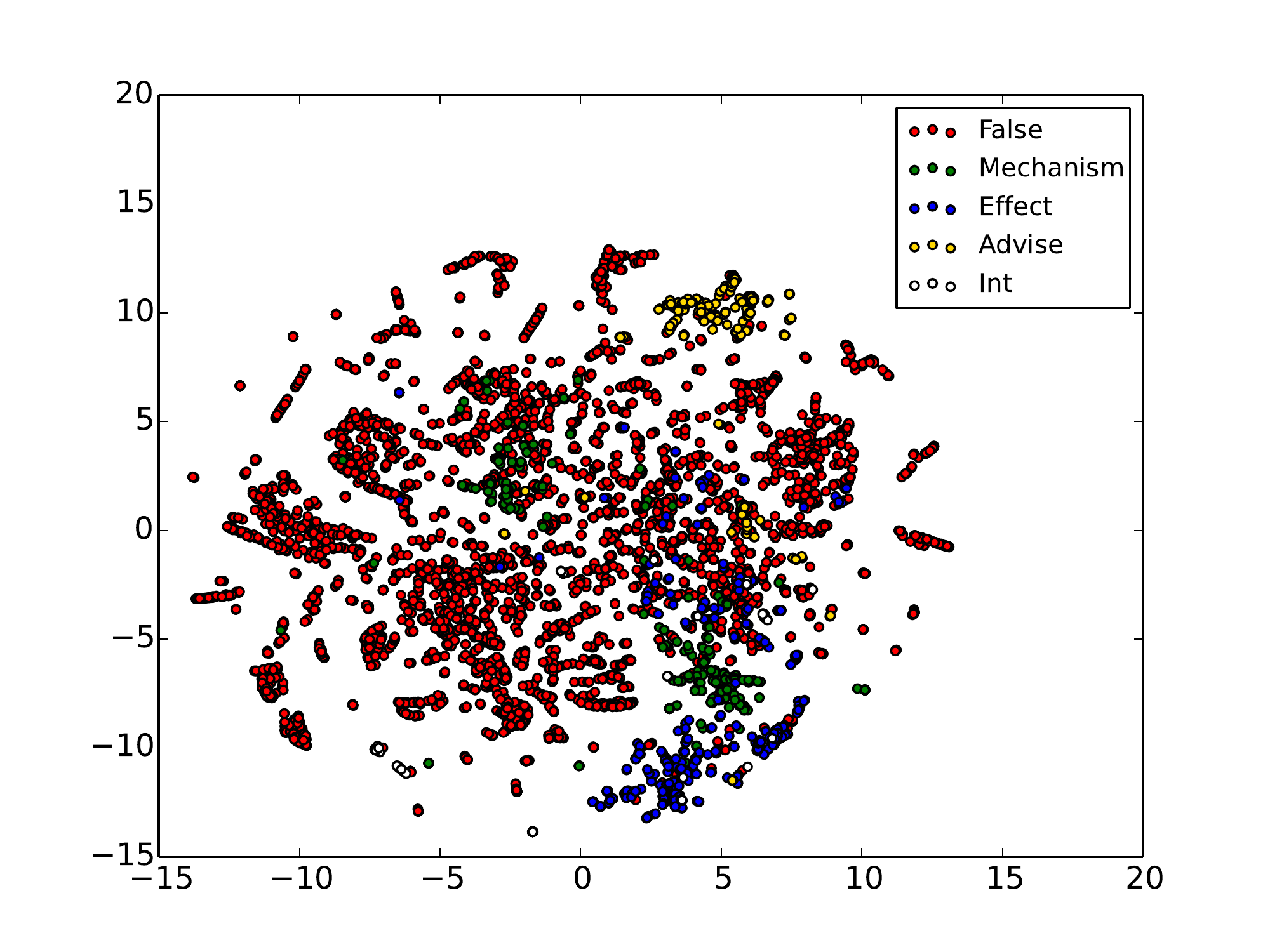}}\label{subfig:features_dynamic_Att}}
\subfigure[Dynamic word embedding + 2ATT] {{\includegraphics[height=1.73in,angle=0]{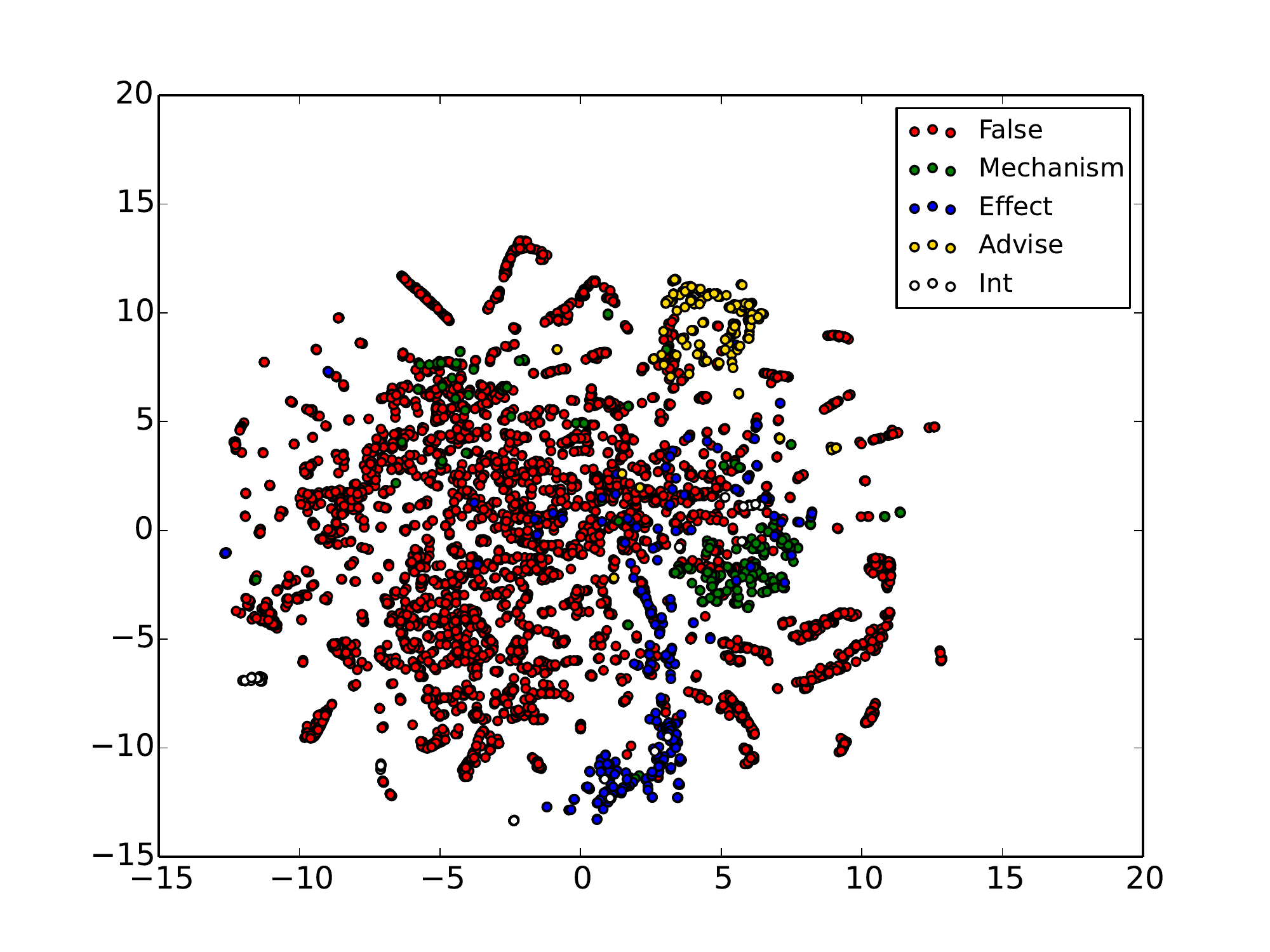}}\label{subfig:features_dynamic_2Att}}
\caption{The features which mapped to 2D}
\label{fig:2Dfeatures}
\end{figure*}

\subsection{Hyperparameter Settings and Training}
We use TensorFlow \cite{abadi2016tensorflow} r0.11 to implement the proposed model. The input of each word is an ordered triple (\textit{word, relative distance from drug1, relative distance from drug2}). The sentence, which is represented as a matrix, is fed to the model. The output of the model is a $|C|$-dimensional vector representing the probabilities of being corresponding DDI. It is the network, parameters, and hyperparameters which decides the output vector. The network's parameters are adjusted during training, where the hyperparameters are tuned by hand. The hyperparameters after tuning are as follows. The word embedding's dimension $d_{WE}=100$, the position embedding's dimension $d_{PE}=10$, the hidden state's dimension $d_h=230$, the probability of dropout $Pr_d=0.5$, other hyperparameters which are not shown here are set to TensorFlow's default values.

The word embedding is initialized by pre-trained word vectors using GloVe \cite{Pennington2014Glove}, while other parameters are initialized randomly. During each training step, a mini-batch (the mini-batch size $|\mathcal M|=60$ in our implementation) of sentences is selected from training set. The gradient of objective function is calculated for parameters updating (See Section \ref{subsec:cla_train}). 

\begin{table}[ht]
\caption{Prediction results}
\label{tab:prediction}
\centering
\begin{tabular}{ccccccc}
 \hline
 \multirow{1}{*}{\tabincell{c}{}}  & \multicolumn{5}{c}{Classified as} & \multirow{2}{*}{Sum}\\
 \cline{2-6}
   ~ & False & Mechanism & Effect & Advise & Int & ~\\
 \hline
 \hline
{False} & 4490 & 138 & 49 & 45 & 15 & 4737\\
 \hline
 {Mechanism} & 68 & 229 & 2 & 3 & 0 & 302\\
 \hline
 {Effect} & 101 & 12 & 230 & 15 & 2 & 360\\
 \hline
 {Advise} & 49 & 5 & 0 & 165 & 2 & 221\\
  \hline
 {Int} & 13 & 3 & 37 & 0 & 43 & 96\\
   \hline
 {Sum} & 4721 & 387 & 318 & 228 & 62 & 5716\\
 \hline
 \end{tabular}
\end{table}

Figure \ref{fig:train_process} shows the training process. The objective function $J(\theta)$ is declining as the training mini-batches continuously sent to the model. As the testing mini-batches, the $J(\theta)$ function is fluctuating while its overall trend is descending. The instances in testing set are not participated in training so that $J(\theta)$ function is not descending so fast. However, training and testing instances have similar distribution in sample space, causing that testing instances' $J(\theta)$ tends to be smaller along with the training process. $J(\theta)$ has inverse relationship with the performance measurement. The F1 score is getting fluctuating around a specific value after enough training steps. The reason why fluctuating range is considerable is that only a tiny part of the whole training or testing set has been calculated the F1 score. Testing the whole set during every step is time consuming and not necessary. We will evaluate the model on the whole testing set in Section \ref{subsec:expresult}.

\subsection{Experimental Results}\label{subsec:expresult}

We save our model every 100 step and predict all the DDIs of the instances in the testing set. These predictions' F1 score is shown in figure \ref{fig:testf1}. To demonstrate the sentence level attention layer is effective, we drop this layer and then directly use $h^*$ for softmax classification (See figure \ref{fig:arch}). The result is shown with ``RNN + dynamic word embedding + ATT'' curve, which illustrates that the sentence level attention layer contributes to a more accurate model.

Whether a dynamic or static word embedding is better for a DDI extraction task is under consideration. Nguyen \emph{et al.} \cite{Nguyen2015Relation} shows that updating word embedding at the time of other parameters being trained makes a better performance in relation extraction task. We let the embedding be static when training, while other conditions are all the same. The ``RNN + static word embedding + 2ATT'' curve shows this case. We can draw a conclusion that updating the initialized word embedding trains more suitable word vectors for the task, which promotes the performance.

We compare our best F1 score with other state-of-the-art approaches in table \ref{tab:performance}, which shows our model has competitive advantage in dealing with drug-drug interaction extraction. The predictions confusion matrix is shown in table \ref{tab:prediction}. The DDIs other than false being classified as false makes most of the classification error. It may perform better if a classifier which can tells true and false DDI apart is trained. We leave this two-stage classifier to our future work. Another phenomenon is that the ``Int'' type is often classified as ``Effect''. The ``Int'' sentence describes there exists interaction between two drugs and this information implies the two drugs' combination will have good or bed effect. That's the reason why ``Int'' and ``Effect'' are often obfuscated. 

To evaluate the features our model captured, we employ scikit-learn\cite{scikit-learn}'s  t-SNE class \footnote{http://scikit-learn.org/stable/modules/generated/sklearn.manifold.TSNE.html} to map high dimensional feature vectors to 2-dimensional vectors, which can be depicted on a plane. We depict all the features of the instances in testing set, as shown in figure \ref{fig:2Dfeatures}. The RNN model using dynamic word embedding and 2 layers of attention is the most distinguishable one. Unfortunately, the classifier can not classify all the instances into correct classes. Comparing table \ref{tab:prediction} with figure \ref{subfig:features_dynamic_2Att}, both of which are from the best performed model, we can observe some conclusions. The ``Int'' DDIs are often misclassified as ``Effect'', for the reason that some of the ``Int'' points are in the ``Effect'' cluster. The ``Effect'' points are too scattered so that plenty of ``Effect'' DDIs are classified to other types. The ``Mechanism'' points are gathered around two clusters, causing that most of the ``mechanism'' DDIs are classified to two types: ``False'' and ``Mechanism''. In short, the visualizability of feature mapping gives better explanations for the prediction results and the quality of captured features.

\section{Conclusion and Future Work}
To conclude, we propose a recurrent neural network with multiple attention layers to extract DDIs from biomedical text. The sentence level attention layer, which combines other sentences containing the same drugs, has been added to our model. The experiments shows that our model outperforms the state-of-the-art DDI extraction systems. Task relevant word embedding and two attention layers improved the performance to some extent.

The imbalance of the classes and the ambiguity of semantics cause most of the misclassifications. We consider that instance generation using generative adversarial networks would cover the instance shortage in specific category. It is also reasonable to use distant supervision learning (which utilize other relevant material) for knowledge supplement and obtain a better performed DDI extraction system.

\section*{Acknowledgment}

This work is supported by the NSFC under Grant 61303191, 61303190, 61402504, 61103015.

\balance
\bibliographystyle{manuscript}

\bibliography{cite}

\end{document}